\documentclass[lettersize,journal]{IEEEtran}
\usepackage{amsmath,amsfonts}
\usepackage{algorithmic}
\usepackage{algorithm}
\usepackage{array}
\usepackage[caption=false,font=normalsize,labelfont=sf,textfont=sf]{subfig}
\usepackage{textcomp}
\usepackage{stfloats}
\usepackage{url}
\usepackage{verbatim}
\usepackage{graphicx}
\usepackage{cite}
\usepackage{times}
\usepackage{soul}
\usepackage{url}
\usepackage[hidelinks]{hyperref}
\usepackage[utf8]{inputenc}
\usepackage[small]{caption}
\usepackage{graphicx}
\usepackage{amsmath}
\usepackage{amsthm}
\usepackage{booktabs}
\usepackage{algorithm}
\usepackage{algorithmic}
\usepackage[switch]{lineno}
\usepackage{amssymb} 
\usepackage{svg}
\usepackage{epstopdf}
\usepackage{colortbl}
\usepackage{hhline}
\usepackage{stfloats}
\usepackage{float}
\usepackage{xspace}
\usepackage{setspace}
\usepackage{overpic}
\usepackage{bbding}
\usepackage{multirow}
\usepackage{CJKutf8}
\usepackage{float}

\begin{document}

\title{Point Cloud Completion Guided by Prior Knowledge via Causal Inference}

\author{Songxue Gao, Chuanqi Jiao, Ruidong Chen, Weijie Wang*, Weizhi Nie}

\markboth{Journal of \LaTeX\ Class Files,~Vol.~14, No.~8, August~2021}%
{Shell \MakeLowercase{\textit{et al.}}: Point-PC: Point Cloud Completion Guided by Prior Knowledge via Causal Inference}


\maketitle

\begin{abstract}
Point cloud completion aims to recover raw point clouds captured by scanners from partial observations caused by occlusion and limited view angles. This makes it hard to recover details because the global feature is unlikely to capture the full details of all missing parts. In this paper, we propose a novel approach to point cloud completion task called Point-PC, which uses a memory network to retrieve shape priors and designs a causal inference model to filter missing shape information as supplemental geometric information to aid point cloud completion. Specifically, we propose a memory operating mechanism where the complete shape features and the corresponding shapes are stored in the form of ``key-value'' pairs. To retrieve similar shapes from the partial input, we also apply a contrastive learning-based pre-training scheme to transfer the features of incomplete shapes into the domain of complete shape features. Experimental results on the ShapeNet-55, PCN, and KITTI datasets demonstrate that Point-PC outperforms the state-of-the-art methods.
\end{abstract}


\section{Introduction}
\IEEEPARstart{W}{ith} more people using 3D scanners and RGB-D cameras, 3D vision has become one of the most popular topics for research in recent years 
\cite{Han2019ShapeCaptionerGC,Han2018DeepSU,tan2022projected, nie2019hgan}. Among all the 3D descriptors\cite{Wang2018Pixel2MeshG3,Xie2020Pix2VoxMC,Qi2017PointNetDL}, the point cloud stands out because of its remarkable ability to render spatial structure at a lower computational cost. 
However, due to occlusion, view angles, and limitations of sensor resolution, raw point clouds are usually sparse and defective \cite{Wen2021Cycle4CompletionUP,Wen2020PointCC,Wen2022PMPNetPC}. Consequently, point cloud completion becomes essential.

\begin{figure}[ht]
\centering
\includegraphics[width=0.9\linewidth]{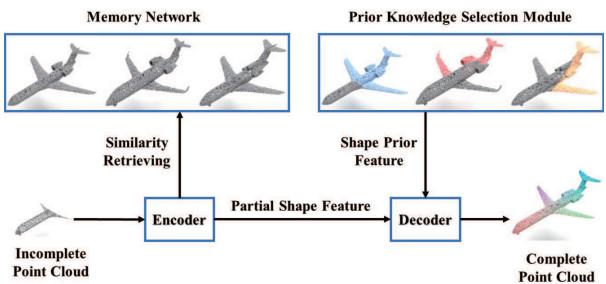}
\caption{Point-PC is proposed for point cloud completion. Point-PC proposes a novel paradigm that finds similar shape information as prior knowledge to help the model handle the point cloud completion problem. Furthermore, our approach also selects geometric information from shape priors (blue, red, and yellow points) guided by causal inference.} 
\label{Figure1}
\vspace{-.4cm}
\end{figure}

Benefiting from large-scale point cloud datasets \cite{Chang2015ShapeNetAI,Yuan2018PCNPC,Geiger2013VisionMR}, massive efficient learning-based methods for point cloud completion have emerged. The pioneering work is PCN \cite{Yuan2018PCNPC} which encoded the input shape into a global feature and decoded it using a folding operation. Following an encoder-decoder pattern, several methods such as NSFA \cite{Zhang2020DetailPP} and GRNet \cite{Xie2020GRNetGR} have emerged.
Subsequent research has prioritized enhancing the decoding aspect of generating point clouds with detailed geometric structures.
SA-Net \cite{Wen2020PointCC} and PFNet \cite{Huang2020PFNetPF} increased the density of point clouds hierarchically. Such a coarse-to-fine pattern achieves better performance since more constraints are imposed on the generation process.
 
Most recent methods incorporate geometry-aware modules into a transformer-based structure. 
PoinTr \cite{Yu2021PoinTrDP} employed a KNN model to enable transformers, thereby improving the ability to utilize the inductive bias related to 3D geometric structures.
SeedFormer \cite{Zhou2022SeedFormerPS} aggregated neighboring points by the proposed Upsample Transformer to incorporate valuable local information into the generation process.
These two methods formulate the point cloud completion task as a set-to-set translation task, where complex dependency is learned among the point groups. 
While studies in this field have demonstrated intriguing performance achievements,
most existing methods \cite{Wang2022PointAttNYO,Yan2022FBNetFN} have always failed to jump out of the paradigm where the query points are generated from a global feature vector extracted from partial inputs in the first place, 
followed by an upsampling module to increase the density of the output.
Many approaches used the same framework to handle the point cloud completion problem \cite{Zhang2022PartialtoPartialPG,Cao2022PointCC}. However, there are two drawbacks to the paradigm: 
1) An incomplete shape makes it hard to learn detailed structure information and build a clear relationship between the complete point cloud model; 
2) A global feature like this is spread out and does not keep much fine-grained information for the up-sample phase. Because of this, geometry-aware models can not learn complex structures if they know less about geometry. 

To tackle these challenges, our insight into revealing detailed geometry is inspired by human memory. We often exploit fragmented pieces of cues to retrieve similar 3D shapes from our memories and let the shape priors guide the reconstruction. Based on this, we propose a new memory-based framework for completing point clouds (Point-PC). This framework uses a memory network to get shape priors and an effective causal inference model to choose missing shape information as additional geometric information to help complete point clouds.
Thus the framework consists of four components: a partial shape encoder, the memory network, the prior knowledge selection module, and the shape decoder. 
Compared with the above-mentioned methods, our proposed method takes one step forward to introduce shape prior explicitly from an external knowledge base. In this way, the geometry-aware structure is enabled to precisely capture the fine-grained details and geometric relations between neighboring points. Even encountering a large unseen points ratio, complete shapes generated by our method still meet the requirements of smooth surfaces, neat edges, and sharp corners. 
First, we construct an operating strategy to store, write, and read the memory. Specifically, we store the memory in a ``key-value'' pair. The key can be updated according to the similarity between the value and the corresponding ground truth. 
This writing strategy associates the value with incomplete inputs from multiple perspectives and enhances the memory network's geometry perception of incomplete shapes. 
To retrieve the precise shapes, we design a partial-complete pre-training scheme with two learning mechanisms by contrastive learning to overcome the distribution biases between the two modalities of point clouds: (i) cross-modality learning to align partial features into the complete domain, (ii) intra-modality learning in the partial modality to eliminate the interference from various view angles and incomplete ratios. 
Our partial-complete pre-training aligns partial features to be consistent with complete features and encourages partial feature extraction to be invariant to viewpoints, thus significantly improving the accuracy of shape prior retrieval.
With the help of the pre-training scheme and the backdoor adjustment, Point-PC explicitly leverages shape priors in the memory network to recover the heavily flawed point clouds on complex occasions.
As the retrieved shapes should not change the original structure of the input partial point cloud, to better leverage the retrieved shapes, we also take advantage of causal inference to extract useful information from the retrieved shapes. 
To achieve the best prior knowledge information, we construct a causal graph to analyze the unrelated shape information of prior shapes. 
Specifically, we select the shape of prior features as the confounder and stratify it with do-calculus. After cutting off the backdoor path, the features extracted from the shape priors are optimized, and redundant information is removed.
To leverage information from both partial points and shape priors, the input partial features are concatenated with the output features of the backdoor adjustment module and then fed into a decoder to predict the final 3D complete shape.
The main contributions of our work are as follows:
\begin{itemize}
    \item We propose a novel memory-based 3D point cloud completion network, Point-PC, to supplement geometric information from prior knowledge explicitly.
    \item We introduce causal inference to further refine the shape prior, so as to eliminate the distraction of irrelevant information.
    \item We apply qualitative and quantitative experiments on ShapeNet-55, PCN, and KITTI datasets, which shows that Point-PC improves the accuracy and plausibility of point cloud completion, and outperforms previous SOTA methods.
\end{itemize}

The following provides an overview of the remaining content in this paper: Section 2 covers related work, Section 3 presents the proposed solution, Section 4 describes the conducted experiments along with a summary of the results, and finally, Section 5 concludes this work and outlines potential avenues for future research.

\section{Related Work}
\subsection{Point cloud Completion}
The pioneering deep learning-based approach, PCN \cite{Yuan2018PCNPC}, generates a preliminary completion by utilizing a learned global feature and subsequently applies upsampling, assuming that a 3D object resides on a 2D manifold.
Later research focuses on mitigating mature learning-based structures. Some previous methods \cite{Liu2019PointVoxelCF,Liu2019RelationShapeCN} voxelized the point cloud into binary voxels to migrate 3D convolutions, which cubically increased the computational cost, whereas other methods \cite{Huang2020PFNetPF,Mandikal2019Dense3P} process coordinates directly by Multi-Layer Perceptrons, yet loses geometric information with pooling-based aggregation operations. These two kinds of completion methods ignore relation and context across points, thus failing to preserve regional information of local patterns. To solve this problem, TopNet \cite{Tchapmi2019TopNetSP} constrains the point completion process as the growth of a hierarchical rooted tree where several child points are projected by a parent point in a feature expansion layer. On the other hand, SnowflakeNet \cite{Xiang2021SnowflakeNetPC} models point cloud completion procedure as the generation of a snowflake. 
Most recent state-of-the-art completion methods focus on the decoding process of recovering fine details instead of providing sufficient geometric guidance from partial inputs in the encoding process. By breaking the point cloud into several sequential patches, transformer-based methods \cite{Guo2021PCTPC,Yu2021PoinTrDP,Zhou2022SeedFormerPS} are proved to efficiently handle large-scale point cloud and enhance relations between neighboring points, which outperform and dominate the research prospect. Nevertheless, upsample and expansion modules among the aforementioned methods are based on a global feature vector due to its simplicity, which prevents them from precisely capturing the detailed geometries and structures of 3D shapes, therefore it is unable for these methods to arrange the well-structured point splitting into local regions. In order to address this problem by integrating more geometric information explicitly, we utilize a memory network to provide rich structural details and enhance neighboring relations to recover local regions.

\begin{figure*}
\centering
\begin{overpic}[width=\linewidth]{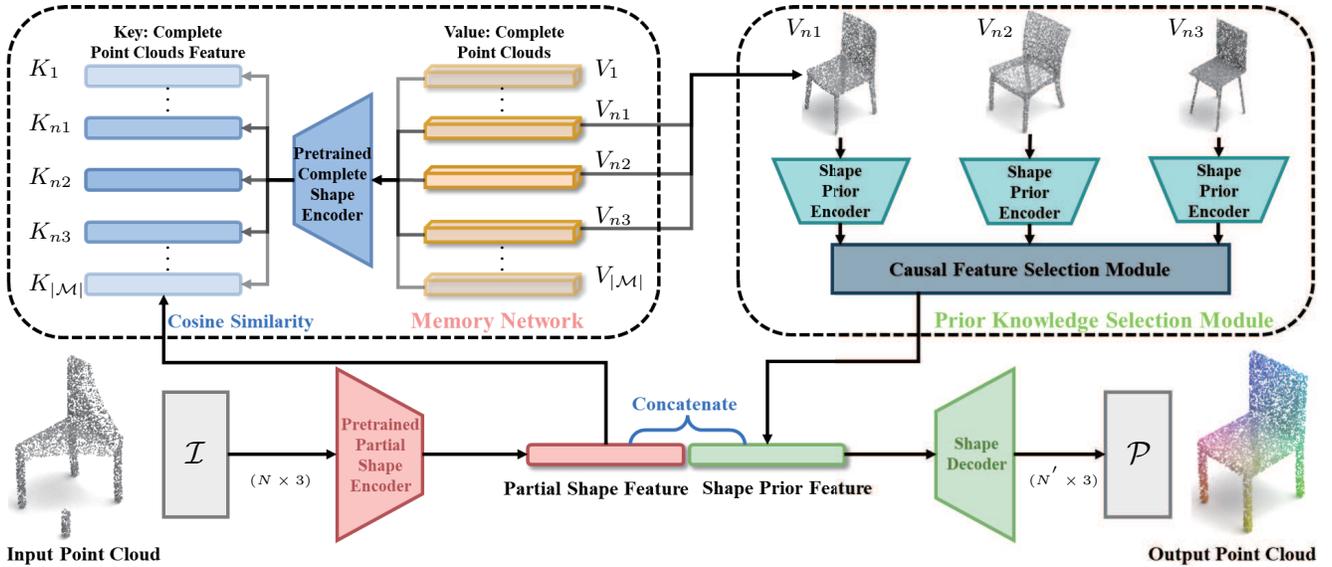}
\put(3.5,38){\small{$K_{1}$}}
\put(13.5,35.4){\small{$\vdots$}}
\put(3.5,34){\small{$K_{n1}$}}
\put(3.5,30){\small{$K_{n2}$}}
\put(3.5,26){\small{$K_{n3}$}}
\put(13.5,23.7){\small{$\vdots$}}
\put(3.5,22.2){\small{$K_{|\mathcal{M}|}$}}
\put(45,38){\small{$V_{1}$}}
\put(38,35.4){\small{$\vdots$}}
\put(45,35){\small{$V_{n1}$}}
\put(45,31.3){\small{$V_{n2}$}}
\put(45,27.2){\small{$V_{n3}$}}
\put(38,23.7){\small{$\vdots$}}
\put(45,22.8){\small{$V_{|\mathcal{M}|}$}}
\put(59,41){\small{$V_{n1}$}}
\put(73,41){\small{$V_{n2}$}}
\put(87,41){\small{$V_{n3}$}}
\put(15,9.5){\large{$\mathcal{I}$}}
\put(19.7,8){\tiny{$(N\times3)$}}
\put(84,9.5){\large{$\mathcal{P}$}}
\put(76.8,8){\tiny{$(N^{'}\times3)$}}
\end{overpic}
\caption{The overall architecture of Point-PC, which consists of four main modules: (i) pre-trained partial shape encoder, (ii) memory network, (iii) prior knowledge selection module, and (iv) shape decoder. The pre-trained encoder extracts features from the partial input, which is then fed into memory. The memory network retrieves shape priors with sufficient geometric information. Moreover, the prior knowledge selection module selects useful information from the prior shapes. The shape decoder takes the concatenation of the partial shape feature and the shape prior feature to generate the complete point cloud.}
\label{Figure2}
\end{figure*}

\subsection{Memory Network}
The Memory Network \cite{Weston2015MemoryN} was initially presented in dialog systems to save scene information and realize the functionality of long-term memory. However, the original design of the Memory Network just vectorizes and saves the original text without proper modification, thus limiting the promotion of the model. Further works \cite{Liu2017GatedEM} reinforce the Memory Network so that it can be trained in an end-to-end way. Hierarchical Memory Network \cite{Chandar2016HierarchicalMN} stores and searches memory in a hierarchical structure to speed up calculations when implementing large-scale memory. Key-Value Memory Network \cite{Miller2016KeyValueMN} stores memory slots in a ``key-value" pair where the key module is responsible for scoring the degree of correlation between memory and queries, while the value module is responsible for weighting and summing the values of the memory to obtain the output. In our work, we further extend the application of ``key-value" structured memory into point cloud completion and reveal its ability to preserve high-quality geometry details through a well-designed pre-training method.

\subsection{Self-supervised Pre-training}
To learn informative representations from invariant 3D features,
Many previous studies have utilized self-supervised pre-training techniques, incorporating views of the same scene, diverse modalities such as depth and RGB images, or even distinct formats like point clouds and voxels. Contrastive Scene Contexts \cite{Hou2020ExploringD3} divides the point cloud of a scene into multiple regions and applies contrastive learning individually within each region. This approach effectively utilizes both point-level correspondences and spatial contexts within the scene.
DepthContrast \cite{Zhang2021SelfSupervisedPO} works with single-view depth scans for self-supervised learning and extends the successful MoCo \cite{Chen2020ImprovedBW} pipeline to jointly train with point and voxel input formats. CLIP2POINT \cite{Huang2022CLIP2PointTC} proposes an image-depth pre-training method to align multi-view depth features with CLIP visual features so as to transfer CLIP knowledge to the 3D category-level discrimination and adapt it to point cloud classification. We design a partial-complete pre-training scheme by contrastive learning to discriminate the modality difference between partial and complete point clouds. By this means, Point-PC learns fine-grained information association and retrieves similar complete shapes based on the partial point cloud. Nonetheless, we further leverage causal inference to refine the redundant shape priors.

\subsection{Causal Inference}
The causal inference was first introduced by \cite{Pearl2000CausalityMR}. Recent research \cite{Hu2021DistillingCE} has shown that causal inference is beneficial to various fields in computer vision. VC R-CNN \cite{Wang2020VisualCR} proposes that observational bias causes the model to tend to make predictions based on co-occurrence information while ignoring some common-sense causal relationships, and attempts to extract a visual feature that contains common sense through causal intervention. 
CONTA \cite{Zhang2020CausalIF} ascribes the uncertain boundaries of pseudo-masks to confounding context and employs backdoor adjustment to remove the confounding factors. This approach generates improved pixel-level pseudo masks using only image-level labels.
Ifsl \cite{Yue2020InterventionalFL} posits that pre-trained knowledge constitutes a confounding factor leading to spurious correlations between sample features and class labels in the support set and employs backdoor adjustment to eliminate this bias.
To the best of our knowledge, we are the first to introduce causal inference to point cloud completion. 
Our approach incorporates a causal feature fusion strategy aimed at mitigating the confounding effect in shape priors. This strategy directs the decoder to focus on causal features, leading to greater robustness in the memory network.

\section{Our Approach}
Figure.\ref{Figure2} shows the framework of Point-PC, which includes four main parts: 1) Pre-train encoder: it is used to extract the partial shape's feature vector; 2) Memory network: it is used to learn the similarity between partial shape and corresponding complete shape; 3) Prior knowledge selection module: it is used to select the useful information from prior complete shapes and help the partial shape complete the missing structural information. We will detail these modules in the next subsections; 4) Shape decoder: it is used to accept the concatenated features and predict the complete point cloud. We will detail each of our designs in the following.

\subsection{Memory Network}
The memory network aims to learn the dependency of partial and complete shapes in feature space and produce the prior shapes.
Denote the input set of partial point clouds as $S = {\left\{\mathcal{I}_{i}\right\}}_{i=1}^{|S|}$, where $\mathcal{I}_{i} \in \mathbb{R}^{N \times 3}$ represents each point in the object, and $N$ is the point number of a shape. We construct the memory network in a ``key-value-age'' formation. The ``key'' and ``value'' represent complete shape features and the corresponding 3D shapes, respectively. The ``age'' indicates how long the corresponding ``key-value'' pair has been established. Therefore, the memory item is denoted as $\mathcal{M}={(K_i,V_i,A_i)_{i=1}^{|\mathcal{M}|}}$, where $|\mathcal{M}|$ is the size of the memory. 

Compared with other methods, the memory network utilizes the ``key'' and ``value'' to improve the effectiveness of prior shapes. Meanwhile, the ``key'' and ``value'' can also be updated by the training data and improve the relevance of obtaining prior information. Next, we will introduce the model update and retrieval process in two parts.

\subsubsection{Update Strategy}
$K_i$ is extracted through the pre-trained complete shape encoder from $V_i$, which can be denoted as $F^{V_i}$. It is worth noting that the updating strategy only works at training because we take the training set as our external knowledge base, which can not be available during testing. 

We compute the cosine similarity between $F^{\mathcal{I}}$ and $F^{V_i}$ to match a ``key-value'' pair as follows:
\begin{equation}
    \label{eq1}
    Sim_{key}\left(F^{\mathcal{I}}, F^{V_i}\right)=\frac{F^{\mathcal{I}} \cdot F^{V_i}}{\|F^{\mathcal{I}}\|\left\|F^{V_i}\right\|}.
\end{equation}

To measure whether it is a valid match, we adopt the Chamfer distance \cite{Yuan2018PCNPC} as the similarity measurement between the corresponding ground truth $\mathcal{V}$ and the value $V_{i}$ in 3D space. If the Chamfer Distance $Sim_{value}$ is lower than a threshold $\delta$ (discussed in Section 4.6), it is a positive match and vice versa. For a positive match, the value $V_{n_0}$ stays unchanged, while the key $F^{V_{n_0}}$ is updated as below:
\begin{equation}
    \label{eq2}
    F^{V_{n_0}}=\frac{F^{\mathcal{I}}+F^{V_{n_0}}}{\left\|F^{\mathcal{I}}+F^{V_{n_1}}\right\|},
\end{equation}
where $n_{0}=\arg \max _{i} Sim_{key}\left(F^{\mathcal{I}}, F^{V_i}\right)$. In the meantime, except for the corresponding age $A_{n_{0}}$ to be set to zero, all the other ages should be increased by one.
For a negative match, $\mathcal{V}$ is read into the memory and should overwrite the oldest slot as follows:
\begin{equation}
    \label{eq3}
    K_{n_1} = F^{\mathcal{I}}, V_{n_1} = \mathcal{V},
\end{equation}
where $n_{1}$ depends on $n_{1} = \arg \max _{i}\left(A_{i}\right)$. The ages here are updated in the same way as mentioned above. In this way, the memory network reinforces its reception ability with similar shapes, saves the unknown shapes, and refreshes the oldest memory slot.

\begin{algorithm}[tb]
    \caption{Update and Query Strategy}
    \label{algo1}
    \textbf{Input}: partial point cloud feature $F^{\mathcal{I}}$\\
    \textbf{Hyper-parameter}: similarity threshold $\delta$\\
    \textbf{Output}: shape priors $V_{n_i}$
    \begin{algorithmic}[1] 
        \STATE Let $i=0$.
        \WHILE {$i \leq |\mathcal{M}|-1$}
        \STATE Compute $Sim_{key}\left(F^{\mathcal{I}}, F^{V_i}\right)$ by Eq.~\ref{eq1}.
        \IF {$(Sim_{value}\left(\mathcal{V}, V_{n_1}\right) \geq \delta)$}
        \STATE Let $n_{0}=\arg \max _{i} Sim_{key}\left(F^{\mathcal{I}}, F^{V_i}\right)$,
        \STATE Update $K_{n_0}$ by Eq.~\ref{eq2},
        \STATE Set $A_{n_{0}}=0, A_{i}=A_{i}+1\left(i \ne n_0 \right)$.
        \ELSE
        \STATE Let $n_{1} = \arg \max _{i}\left(A_{i}\right)$.
        \STATE Update $K_{n_1}$ and $V_{n_1}$ by Eq.~\ref{eq3},
        \STATE Set $A_{n_1} = 0, A_{i} = A_{i}+1\left(i \ne n_1 \right)$.
        \ENDIF
        \ENDWHILE
        \STATE \textbf{return} $V_{n_i}$ by Eq.~\ref{eq4}.
    \end{algorithmic}
\end{algorithm}

\subsubsection{Query Strategy}
We propose a query strategy for obtaining shape priors that are rich in geometric information for completion and very similar to the partial input. These shape priors are the values in the memory, which are complete point clouds. To fix the number of shape priors fed forward, we retrieve $\hat{k}$ shapes through top-$\hat{k}$ keys with the largest similarity for convenience, which can be formulated as:
\begin{equation}
\label{eq4}
    V=\left[V_{n_i}| n_i=\arg\max _{i} Sim_{key}\left(F^{\mathcal{I}}, F^{V_{i}}\right)\right].
\end{equation}

For a more organized elaboration of the update and query process, we describe the simplified procedure of the memory network in Algorithm \ref{algo1}.

\subsection{Pre-training Scheme}
\begin{figure}[ht]
\begin{overpic}[width=\linewidth]{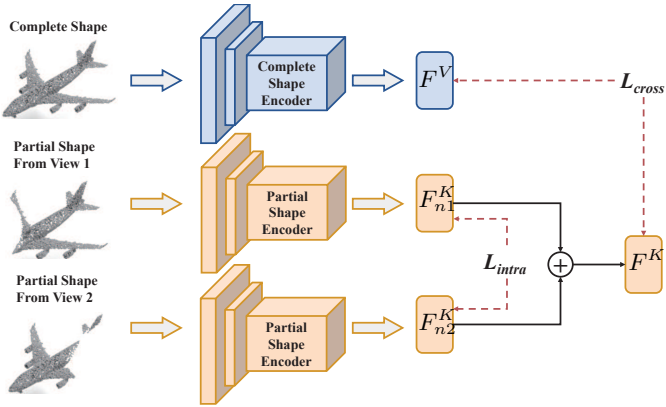}
\put(63,51){\small{$F^{V}$}}
\put(63,33){\small{$F^{K}_{n1}$}}
\put(63,14.5){\small{$F^{K}_{n2}$}}
\put(94.5,23.5){\small{$F^{K}$}}
\end{overpic}
\caption{Architecture of contrastive learning-based pre-training scheme. We adopt one NT-Xent loss between pairs of partial shape features extracted from the partial shape encoder and the other between complete shape features and average partial shape features.} 
\label{Figure3}
\end{figure}

The pre-training scheme aims to minimize the distance between partial point clouds and complete point clouds, as well as enhance the consistency of partial shape features.
Given the complete shape denoted as $\mathcal{S}_{i} \in \mathbb{R}^{N \times 3}$, where $N$ is the number of points, we render the corresponding partial ones $\mathcal{I}_{i,n_1}$ and $\mathcal{I}_{i,n_2}$ in different viewpoints and crop different numbers of $n_1$ and $n_2$ points. The overall architecture of the contrastive learning-based pre-training scheme is illustrated in Figure \ref{Figure3}.

\subsubsection{Intra-modality Learning}
Suppose that $\mathcal{I}_{i,n_1}$ and $\mathcal{I}_{i,n_2}$ are fed into the partial shape encoder $E_K$ to extract features $F_{i,n_1}^K,F_{i,n_2}^K \in \mathbb{R}^{1 \times C}$, where $C$ is the feature dimension.
Following the NT-Xent loss in SimCLR \cite{Chen2020ASF}, given a positive pair ($F_{i,n_1}^K$, $F_{i,n_2}^K$), we treat the other $2(N-1)$ examples within a minibatch as negative examples, where $N$ is the size of the minibatch. The intra-modality contrastive loss $\mathcal{L}_{intra}$ can be formulated as:

\begin{equation}
    l_{\text {intra}}\left(i;n_{1},n_{2}\right) =-\log \frac{Sim_{pos}(i;n_{1},n_{2})}{Sim_{neg}(i;n_{1},n_{2})},
\end{equation}
\begin{equation}
    \mathcal{L}_{\text {intra}} =\frac{1}{2N} \sum_{i=1}^{N}\left(l_{\text {intra}}\left(i;n_{1},n_{2}\right)+l_{\text {intra}}\left(i;n_{2},n_{1}\right)\right),
\label{eqintra}
\end{equation}
where $Sim_{pos}(i;n_{1},n_{2})$ and $Sim_{neg}(i;n_{1},n_{2})$ represent the positive and negative cosine similarity between the same partial inputs but with a different incomplete pattern. The cosine similarity function is defined as follows:
\begin{equation}
     \resizebox{.8\linewidth}{!}{$
                \displaystyle
                \begin{aligned}
                    Sim_{pos}(i;n_{1},n_{2}) &= \exp \left(sim\left(F_{i, n_{1}}^{K}, F_{i, n_{2}}^{K}\right) / \tau\right), \\
                    Sim_{neg}(i;n_{1},n_{2}) &= \sum_{j=1}^{N} \mathbb{I}_{[j \neq i]} \exp \left(sim\left(F_{i, n_{1}}^{K}, F_{j, n_{1}}^{K}\right) / \tau\right) \\
                    &+ \sum_{j=1}^{N} \exp \left(sim\left(F_{i, n_{1}}^{K}, F_{j, n_{2}}^{K}\right) / \tau\right), \\
                \end{aligned}
    $}
\end{equation}
where $\mathbb{I}_{[j \neq i]} \in \{0,1\}$ is an indicator function evaluating to 1 if $j \neq i$ and $\tau$ is the temperature parameter which we set to 0.1.

\subsubsection{Cross-modality Learning}
Considering that the partial shape features should remain consistent with the complete shape features, for each ${\mathcal{S}_i}$, we extract features $F_{i}^V \in \mathbb{R}^{1 \times C}$ by the complete shape encoder $E_V$. Together with the partial shape features $F_{i}^K$, the cross-modality contrastive loss $\mathcal{L}_{cross}$ is indicated as follows:
\begin{equation}
    l_{\text {cross}}\left(i;K,V\right) =-\log \frac{Sim_{pos}(i;K,V)}{Sim_{neg}(i;K,V)},
\end{equation}
\begin{equation}
    \mathcal{L}_{\text {cross}}=\frac{1}{2N} \sum_{i=1}^{N}\left(l_{\text {cross}}(i;K,V)+l_{\text {cross}}(i;V,K)\right)
\label{eqcross}
\end{equation}
where $Sim_{pos}(i;K,V)$ and $Sim_{neg}(i;K,V)$ represent the positive and negative cosine similarity between the partial and complete shape features.
The overall pre-training loss function $\mathcal{L}_{pre}$ is the sum of the intra-modality and cross-modality loss $\mathcal{L}_{pre}=\mathcal{L}_{intra}+\mathcal{L}_{cross}$.

\subsection{Prior Knowledge Selection Module}
\begin{figure}[t]
\includegraphics[width=\linewidth]{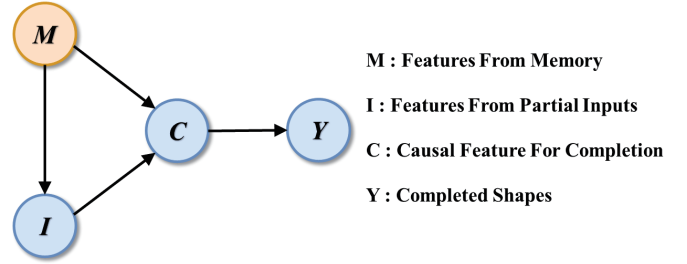}
\caption{Causal graph for Causal Feature Selection Module. Circles represent variables, and arrows represent causal relationships from one variable to another.} \label{Figure4}
\end{figure}

We exploit causal theory \cite{Pearl2013InterpretationAI} to dig out the true causality of the extracted features and generated 3D shapes. The causal graph is shown as Figure \ref{Figure4}.

We list the following explanations for the causalities among the four variables shown in Figure \ref{Figure4}:
\begin{itemize}
    \item $M \rightarrow I $. Since the retrieved shapes share the same semantic structures as the partial inputs, this causal effect is naturally established.
    \item $I \rightarrow C \leftarrow M$. The variable $C$ denotes the causal feature that is truly responsible for the completion result. We not only keep the original part $I$ but also add $M$ as the supplementary information.
    \item $C \rightarrow Y $. The causality reflects the intrinsic association of the feature space and 3D coordinate space.
\end{itemize}

Investigating the causal graph above, we recognize a backdoor path between $M$ and $I$, \textit{i.e.}, $M \rightarrow I $, wherein the $M$ plays a role of confounder between $I$ and $C$. This backdoor path will cause $I$ to create a false correlation with $Y$ even if $I$ is not the only one directly linked to $Y$, resulting in generating low-quality shapes. 
Therefore, it is vital to cut off the backdoor path.

\subsubsection{Backdoor Adjustment}
Instead of modeling the confounded $P(Y|I)$ in Figure \ref{Figure4}, we need to eliminate the backdoor path. According to causal theory, 
we utilize the do-calculus technique on variable $M$ to eliminate the backdoor path by estimating $P_{B}(Y|I)=P(Y|do(I))$ which stratifies the confounder $M$.
We then obtain the following derivations:
\begin{itemize}
    \item The features extracted from memory will not be affected by cutting off the backdoor path. Thus, $P(m)=P_{B}(m)$.
    \item $C$ has nothing to do with the causal effect between the variable $M$ and $I$, which we can get $P_{B}(C|I,m)=P(C|I,m)$.
    \item After the causal intervention, the variable $m$ is independent from $I$, for which we have $P_{B}(m)=P_{B}(m|I)$.
\end{itemize}
$B$ refers to the case when the backdoor path is cut off, and $m \in M$ denotes the confounder sets. Driven by the derivations above, the backdoor adjustment for Figure \ref{Figure4} can be written as:
\begin{equation}
\label{eq_causal}
\resizebox{.7\linewidth}{!}{$
            \displaystyle
    \begin{aligned}
        P(Y\mid{do}(I)) &= \sum_{m \in M}P_{B}(Y|I, m)P_{B}(m|I) \\
        &= \sum_{m \in M}P_{B}(Y|I, m)P_{B}(m) \\
        &= \sum_{m \in M}P(Y|I, m)P(m),
    \end{aligned}
    $}
\end{equation}
where $P(Y|I, m)$ represents the conditional probability given the partial shape feature and confounder; $P(m)$ is the prior probability of the confounder.

\subsubsection{Module Design}
Driven by Eq \ref{eq_causal}, we design the causal feature selection module to alleviate the confounding effect in shape priors.
Our implementation idea is stratifying the confounder and pairing the
partial shape feature with every stratification. 
To achieve this objective, we perform an implicit intervention on feature-wise stratification. Let's consider $\mathcal{H}$ as the index set comprising dimensions of the concatenated shape prior feature obtained from the final layer of the shape prior encoder. We divide $\mathcal{H}$ into n disjoint subsets of equal size. For instance, if the output feature dimension of the shape prior encoder is 384 and we choose the top-3 shape priors with $n = 6$, each subset would consist of a feature dimension index set of size 192. In other words, $\mathcal{H}_{m}$ can be represented as ${192(m-1) + 1, ..., 192m}$.

\begin{itemize}
    \item $P(Y|I, m)=P_{\phi}(Y|cat(F_I, [F_V]_{c}))$, where $F_I$ and $F_V$ are the partial shape feature and the concatenated shape prior feature, respectively. $[F_V]_{c}$ refers to a feature selector that extracts dimensions of $F_V$ based on the index set c. Note that c is defined as $c=\{k|k \in \mathcal{H}_{m} \cap \mathcal{S}_{t}\}$, where $\mathcal{S}_{t}$ represents an index set containing absolute values in $F_V$ exceeding the threshold $t$. Additionally, $\phi$ denotes the parameters of the shape decoder.
    \item $P(m)=1/n$, where We assume that each adjusted feature has an equal prior probability, which is calculated as the reciprocal of the number of confounder sets, represented by $n$. 
\end{itemize}

Thus, the overall feature-wise adjustment is:
\begin{equation}
\label{eq5}
\resizebox{.8\linewidth}{!}{$
            \displaystyle
    P(Y\mid{do}(I))=\frac{1}{n} \sum_{m \in M}P_{\phi}(Y|cat(F_I, [F_V]_{c})).
    $}
\end{equation}

To optimize the $\phi$ in the above Eq \ref{eq5}, we propose a slightly modified L1 Chamfer Distance loss guided by the backdoor adjustment. Let $\mathcal{G}$ be the notation of high-resolution ground truth, and $\mathcal{P}$ be the notation of the completed prediction. The $\mathcal{L}_{caus}$ can be written as:
\begin{equation}
    \mathcal{P} = \Phi(cat(F_I, [F_V]_{c})),
\end{equation}
\begin{equation}
\label{eq6}
    \mathcal{L}_{caus} =\frac{1}{n} \sum_{m \in M}\left(CD-\ell_{1}(\mathcal{P}, \mathcal{G})\right),
\end{equation}
where $\Phi$ represents the shape decoder, and $cat(\cdot, \cdot)$ denotes the concatenate operation. 
The Eq \ref{eq6} encourages the model to generate predictions for the intervened partial-complete probability that remain consistent and stable across various stratification groups, owing to the shared causal features.

We follow the existing works \cite{Yu2021PoinTrDP} to use the L1 Chamfer Distance \cite{Fan2016APS} as a quantitative measurement for the quality of output. Apart from generating $\mathcal{P}$, Point-PC also predicts local centers $\mathcal{C}$ of the completed point cloud. For each prediction, the L1 Chamfer Distance loss function between the central point set and the ground truth $\mathcal{G}$ is calculated as:
\begin{equation}
    \mathcal{L}_{recon} =\frac{1}{|\mathcal{C}|} \sum_{c \in \mathcal{C}} \min _{g \in \mathcal{G}}\|c-g\|_1+\frac{1}{|\mathcal{G}|} \sum_{g \in \mathcal{G}} \min _{c \in \mathcal{C}}\|g-c\|_1.
\end{equation}
The final objective function can be defined as the sum of the losses: $\mathcal{L} = \lambda \mathcal{L}_{caus} + (1-\lambda)\mathcal{L}_{recon}$, where $\lambda$ is a hype-parameter used to control the contribution of different losses in the optimization process.

\section{Experiment}
In this section, we first introduce the implementation details. Then we discuss the evaluation experiments on mainstream benchmarks. We also visualize and analyze the results for both our method and several baseline methods. At last, we provide the ablation study of our method.

\subsection{Results on ShapeNet-55}
\subsubsection{Data.}ShapeNet-55 contains 55 categories of synthetic objects, derived from the ShapeNet dataset. This dataset was first released in PoinTr \cite{Yu2021PoinTrDP}, to improve sample diversity and breadth. They randomly sample 8,192 points to obtain the complete point cloud, with a total of 41,952 point cloud models for training and the rest 10,518 models for testing. The partial point clouds are generated by cutting off certain farthest points from a pre-fixed viewpoint and Keeping 2048 points through the furthest point sample(FPS).
\subsubsection{Quantitative Evaluation.}Following the evaluation setting in \cite{Yu2021PoinTrDP}, 
8 specific viewpoints have been chosen, and the partial point cloud is configured to contain 2,048, 4,096, or 6,144 points, which correspond to 25\%, 50\%, and 75\% of the total points in the entire point cloud, respectively.
In this way, we divide the testing stage into three difficulty degrees simple, moderate, and hard(denoted as CD-S, CD-M, and CD-H).
As shown in Table.~\ref{Table1}, Point-PC achieves 0.83 average CD-$\ell_{2}$ (multiplied by 1000) and 0.479 F-Score@1\% on ShapeNet-55, which demonstrates that Point-PC outperforms the SOTA methods encountering.
Compared with previous methods, Point-PC achieves 0.45, 0.72, and 1.32 CD-$\ell_{2}$ under the three difficulty settings, respectively. The increment of CD-$\ell_{2}$ under CD-M(+0.27) and CD-H(+0.87) strategy also shows that Point-PC better deals with diverse incompleteness levels and diverse incomplete patterns.
Furthermore, we report the results on categories with sufficient(first 5 columns) and insufficient(following 5 columns) training samples. Point-PC performs better than all previous SOTA methods on diverse categories of objects despite the training sample imbalance.
Quantitative results on ShapeNet-55 clearly demonstrate that Point-PC is capable of generating complete point clouds under diverse settings.

\begin{table*}[t]
  \centering
  \resizebox{\textwidth}{!}{
    \begin{tabular}{l|ccccc|ccccc|ccc|cc}
    \toprule
    Methods & Table & Chair &  Airplane & Car   & Sofa  & \multicolumn{1}{p{4.055em}}{Birdhouse} & Bag   & Remote & \multicolumn{1}{p{4.055em}}{Keyboard} & Rocket & CD-S  &  CD-M &  CD-H & \cellcolor[rgb]{ .949,  .949,  .949}CD-Avg $\downarrow$ & \cellcolor[rgb]{ .949,  .949,  .949}F1 $\uparrow$ \\
    \midrule
    FoldingNet & 2.53  & 2.81  & 1.43  & 1.98  & 2.48  & 4.71  & 2.79  & 1.44  & 1.24  & 1.48  & 2.67  & 2.66(-0.01) & 4.05(+1.38) & \cellcolor[rgb]{ .949,  .949,  .949}3.12 & \cellcolor[rgb]{ .949,  .949,  .949}0.082 \\
    PCN   & 2.13  & 2.29  & 1.02  & 1.85  & 2.06  & 4.5   & 2.86  & 1.33  & 0.89  & 1.32  & 1.94  & 1.96(+0.02) & 4.08(+2.14) & \cellcolor[rgb]{ .949,  .949,  .949}2.66 & \cellcolor[rgb]{ .949,  .949,  .949}0.133 \\
    TopNet & 2.21  & 2.53  & 1.14  & 2.18  & 2.36  & 4.83  & 2.93  & 1.49  & 0.95  & 1.32  & 2.26  & 2.16(-0.10) & 4.3(+2.04) & \cellcolor[rgb]{ .949,  .949,  .949}2.91 & \cellcolor[rgb]{ .949,  .949,  .949}0.126 \\
    PFNet & 3.95  & 4.24  & 1.81  & 2.53  & 3.34  & 6.21  & 4.96  & 2.91  & 1.29  & 2.36  & 3.83  & 3.87(+0.04) & 7.97(+4.14) & \cellcolor[rgb]{ .949,  .949,  .949}5.22 & \cellcolor[rgb]{ .949,  .949,  .949}0.339 \\
    GRNet & 1.63  & 1.88  & 1.02  & 1.64  & 1.72  & 2.97  & 2.06  & 1.09  & 0.89  & 1.03  & 1.35  & 1.71(+0.36) & 2.85(+1.5) & \cellcolor[rgb]{ .949,  .949,  .949}1.97 & \cellcolor[rgb]{ .949,  .949,  .949}0.238 \\
    PoinTr & 0.81  & 0.95  & 0.44  & 0.91  & 0.79  & 1.86  & 0.93  & 0.53  & 0.38  & 0.57  & 0.58  & 0.88(+0.30) & 1.79(+1.21) & \cellcolor[rgb]{ .949,  .949,  .949}1.09 & \cellcolor[rgb]{ .949,  .949,  .949}0.464 \\
    SeedFormer & 0.72  & 0.81  & 0.4   & 0.89  & 0.71  & 1.51  & 0.79  & 0.46  & 0.36  & 0.5   & 0.5   & 0.77(+0.27) & 1.49(+0.99) & \cellcolor[rgb]{ .949,  .949,  .949}0.92 & \cellcolor[rgb]{ .949,  .949,  .949}0.472 \\
    \midrule
    Point-PC & \textbf{0.69} & \textbf{0.77} & \textbf{0.38} & \textbf{0.84} & \textbf{0.64} & \textbf{1.37} & \textbf{0.7} & \textbf{0.42} & \textbf{0.31} & \textbf{0.46} & \textbf{0.45} & \textbf{0.72}(+0.27) & \textbf{1.32}(+0.87) & \cellcolor[rgb]{ .949,  .949,  .949}\textbf{0.83} & \cellcolor[rgb]{ .949,  .949,  .949}\textbf{0.479} \\
    \bottomrule
    \end{tabular}}
    \caption{Quantitative results of our methods and several baselines on ShapeNet-55. Detailed results for each method on 10 selected categories are reported, as well as the overall results on 55 categories. CD-S, CD-M, and CD-H represent the CD-$\ell_{2}$ results under the simple, moderate, and hard settings, respectively. Numbers in parentheses represent increments of CD-$\ell_{2}$ results compared to results under the CD-S setting. We show the best results in bold.}
  \label{Table1}%
  \vspace{-4mm}
\end{table*}%

\begin{figure}[ht]
\includegraphics[width=\linewidth]{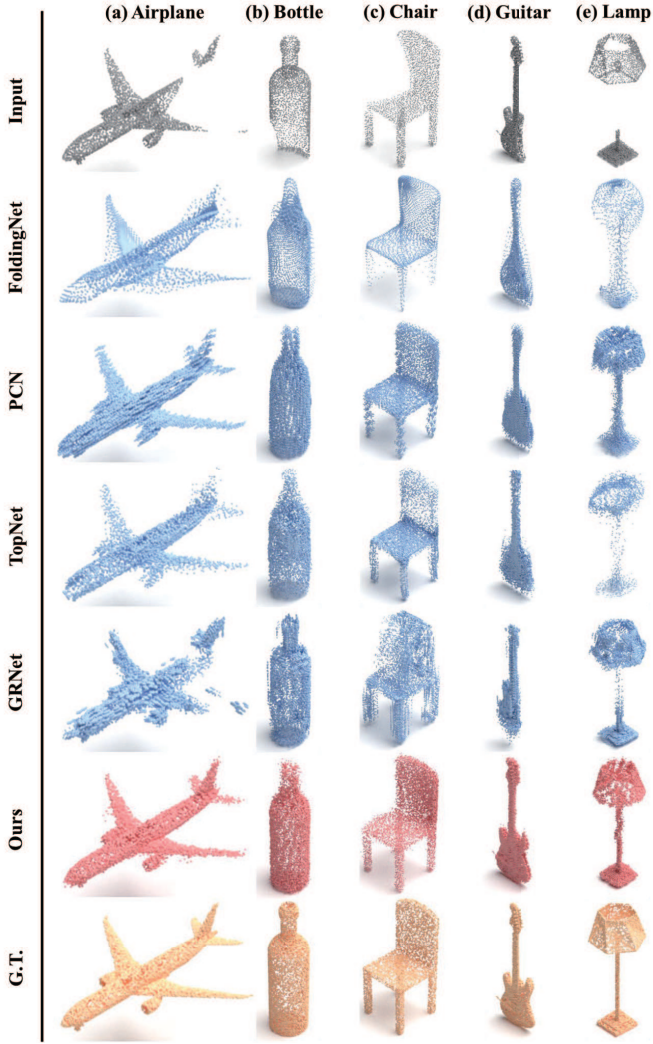}
\caption{Qualitative results on ShapeNet-55 benchmark. 
Each of the methods mentioned above uses the examples in the first row as input to create complete point clouds. Our approach, however, fills in these inputs with a more uniform outline, which distinctly demonstrates the efficacy of our approach.} \label{Figure5}
\vspace{-.4cm}
\end{figure}

\subsubsection{Qualitative Evaluation.}The qualitative comparison results are shown in Figure.~\ref{Figure4}. The proposed Point-PC performs better with fine details than the other methods. 
For example, in the bottle category, Point-PC can predict a smoother and more regular structure of bottle edges compared with the other methods.
Moreover, Point-PC retains the original details of the partial shapes.
In the fifth column of Figure.~\ref{Figure4}, Point-PC not only generates the incomplete lamp bracket with a clear structure but also keeps the texture of the lamp shade, which makes it a more plausible completion than the other methods.
Consequently, Point-PC can effectively learn the geometric information based on the existing partial shape, retrieve similar shape priors based on the learned information and reconstruct complete shapes with more regular arrangements and surface smoothness.

\subsection{Results on ShapeNet-34}
\subsubsection{Data.}
ShapeNet-34 is designed to train on a subset of 34 categories and reserves an additional 21 categories for testing. This setup is to assess how well Point-PC can generalize its performance to new objects from categories that were not encountered during the training stage.
ShapeNet-34 uses a subset of 34 categories for training and leaves 21 unseen categories for testing. We utilize ShapeNet-34 to evaluate the performance of Point-PC on novel objects from categories that do not appear in the training phase.
\subsubsection{Results.}
\begin{figure}[ht]
\small
\includegraphics[width=\linewidth]{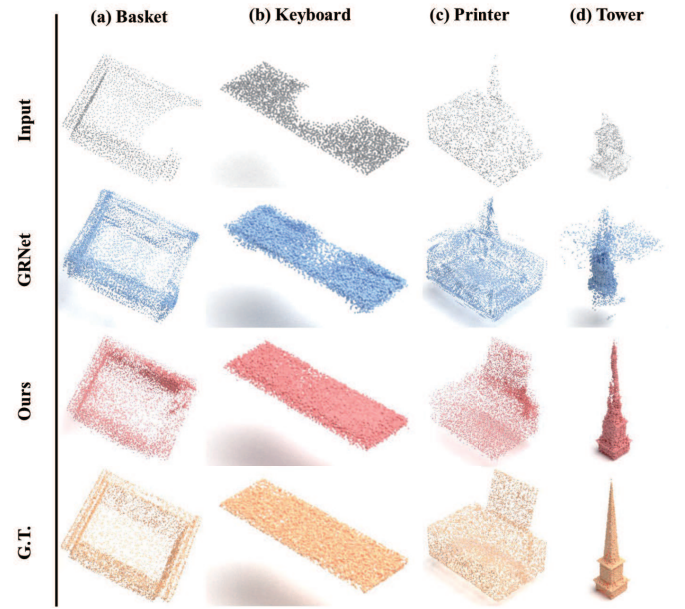}
\caption{
Quantitative results on objects belonging to new categories that were not included in the training dataset. The input partial point cloud, the ground truth, and the predictions made by GRNet and Point-PC are all illustrated.} \label{Figure6}
\vspace{-4mm}
\end{figure}

In Table.~\ref{Table2}, 
we present the metrics for 34 familiar categories and 21 unfamiliar categories under three levels of difficulty evaluated as CD-$\ell_{2}$(multiplied by 1000) and F-Score@1\%.
For Point-PC, we observe fewer gaps between the results of 34 seen categories and 21 unseen categories under each difficulty setting, which demonstrates the superiority of shape priors offered by the memory network.
We also provide the visual comparison with GRNet in Figure.~\ref{Figure6} to show the effectiveness of Point-PC on the unseen categories.

\begin{table}[ht]
  \centering
  \resizebox{\linewidth}{!}{
    \begin{tabular}{l|ccccc|ccccc}
    \toprule
    Methods & \multicolumn{5}{c|}{34 seen categories} & \multicolumn{5}{c}{21 unseen categories} \\
          & CD-S  & CD-M  & CD-H  & \cellcolor[rgb]{ .949,  .949,  .949}CD-Avg & \cellcolor[rgb]{ .949,  .949,  .949}F1 & CD-S  & CD-M  & CD-H  & \cellcolor[rgb]{ .949,  .949,  .949}CD-Avg $\downarrow$ & \cellcolor[rgb]{ .949,  .949,  .949}F1 $\uparrow$ \\
    \midrule
    FoldingNet & 1.86  & 1.81  & 3.38  & \cellcolor[rgb]{ .949,  .949,  .949}2.35 & \cellcolor[rgb]{ .949,  .949,  .949}0.139 & 2.76  & 2.74  & 5.36  & \cellcolor[rgb]{ .949,  .949,  .949}3.62 & \cellcolor[rgb]{ .949,  .949,  .949}0.095 \\
    PCN   & 1..87 & 1.81  & 2.97  & \cellcolor[rgb]{ .949,  .949,  .949}2.22 & \cellcolor[rgb]{ .949,  .949,  .949}0.154 & 3.17  & 3.08  & 5.29  & \cellcolor[rgb]{ .949,  .949,  .949}3.85 & \cellcolor[rgb]{ .949,  .949,  .949}0.101 \\
    TopNet & 1.77  & 1.61  & 3.54  & \cellcolor[rgb]{ .949,  .949,  .949}2.31 & \cellcolor[rgb]{ .949,  .949,  .949}0.171 & 2.62  & 2.43  & 5.44  & \cellcolor[rgb]{ .949,  .949,  .949}3.5 & \cellcolor[rgb]{ .949,  .949,  .949}0.121 \\
    PFNet & 3.16  & 3.19  & 7.71  & \cellcolor[rgb]{ .949,  .949,  .949}4.68 & \cellcolor[rgb]{ .949,  .949,  .949}0.347 & 5.29  & 5.87  & 13.33 & \cellcolor[rgb]{ .949,  .949,  .949}8.16 & \cellcolor[rgb]{ .949,  .949,  .949}0.322 \\
    GRNet & 1.26  & 1.39  & 2.57  & \cellcolor[rgb]{ .949,  .949,  .949}1.74 & \cellcolor[rgb]{ .949,  .949,  .949}0.251 & 1.85  & 2.25  & 4.87  & \cellcolor[rgb]{ .949,  .949,  .949}2.99 & \cellcolor[rgb]{ .949,  .949,  .949}0.216 \\
    PoinTr & 0.76  & 1.05  & 1.88  & \cellcolor[rgb]{ .949,  .949,  .949}1.23 & \cellcolor[rgb]{ .949,  .949,  .949}0.421 & 1.04  & 1.67  & 3.44  & \cellcolor[rgb]{ .949,  .949,  .949}2.05 & \cellcolor[rgb]{ .949,  .949,  .949}0.384 \\
    SeedFormer & 0.48  & 0.7   & 1.3   & \cellcolor[rgb]{ .949,  .949,  .949}0.83 & \cellcolor[rgb]{ .949,  .949,  .949}0.452 & 0.61  & 1.07  & 2.35  & \cellcolor[rgb]{ .949,  .949,  .949}1.34 & \cellcolor[rgb]{ .949,  .949,  .949}0.402 \\
    \midrule
    Point-PC & \textbf{0.42} & 0.55  & \textbf{1.16} & \cellcolor[rgb]{ .949,  .949,  .949}\textbf{0.71} & \cellcolor[rgb]{ .949,  .949,  .949}\textbf{0.464} & \textbf{0.57} & \textbf{0.92} & \textbf{2.09} & \cellcolor[rgb]{ .949,  .949,  .949}\textbf{1.19} & \cellcolor[rgb]{ .949,  .949,  .949}\textbf{0.418} \\
    \bottomrule
    \end{tabular}}
    \caption{Quantitative results on ShapeNet-34 evaluated as CD-$\ell_{2}$(multiplied by 1000) and F-Score@1\%.}
  \label{Table2}%
\end{table}%

\subsection{Results on PCN}
\subsubsection{Data.}
The PCN dataset, a prominent benchmark for point cloud completion, encompasses shapes from 8 different categories. Each shape's ground truth consists of 16,384 points that are uniformly sampled from the mesh surface, while the partial input is formed from 2,048 points derived from 8 randomly chosen viewpoints. To evaluate our method against established benchmarks and to conduct a comparative analysis with leading-edge techniques, we performed experiments on the PCN dataset. We adhered to the standard procedures and evaluation criteria as outlined in \cite{Wang2020CascadedRN,Wen2020PMPNetPC,Xie2020GRNetGR} for our model analysis and ablation studies.

\begin{table}[htbp]
  \centering
  \resizebox{\linewidth}{!}{
    \begin{tabular}{l|cccccccc|c}
    \toprule
    Methods & Air   & Cab   &  Car  &  Cha  &  Lam  &  Sof  & Tab   & Wat   & \cellcolor[rgb]{ .949,  .949,  .949}Avg $\downarrow$ \\
    \midrule
    FoldingNet & 9.49  & 15.8  & 12.61 & 15.55 & 16.41 & 15.97 & 13.65 & 14.99 & \cellcolor[rgb]{ .949,  .949,  .949}14.31 \\
    AtlasNet & 6.37  & 11.94 & 10.1  & 12.06 & 12.37 & 12.99 & 10.33 & 10.61 & \cellcolor[rgb]{ .949,  .949,  .949}10.85 \\
    PCN   & 5.5   & 22.7  & 10.63 & 8.7   & 11    & 11.34 & 11.68 & 8.59  & \cellcolor[rgb]{ .949,  .949,  .949}9.64 \\
    TopNet & 7.61  & 13.31 & 10.9  & 13.82 & 14.44 & 14.78 & 11.22 & 11.12 & \cellcolor[rgb]{ .949,  .949,  .949}12.15 \\
    MSN   & 5.6   & 11.9  & 10.3  & 10.2  & 10.7  & 11.6  & 9.6   & 9.9   & \cellcolor[rgb]{ .949,  .949,  .949}10 \\
    GRNet & 6.45  & 10.37 & 9.45  & 9.41  & 7.96  & 10.51 & 8.44  & 8.04  & \cellcolor[rgb]{ .949,  .949,  .949}8.83 \\
    CRN   & 4.79  & 9.97  & 8.31  & 9.49  & 8.94  & 10.69 & 7.81  & 8.05  & \cellcolor[rgb]{ .949,  .949,  .949}8.51 \\
    NSFA  & 4.76  & 10.18 & 8.63  & 8.53  & 7.03  & 10.53 & 7.35  & 7.48  & \cellcolor[rgb]{ .949,  .949,  .949}8.06 \\
    PMP-Net & 5.65  & 11.24 & 9.64  & 9.51  & 6.95  & 10.83 & 8.72  & 7.25  & \cellcolor[rgb]{ .949,  .949,  .949}8.73 \\
    PoinTr & 4.75  & 10.47 & 8.68  & 9.39  & 7.75  & 10.93 & 7.78  & 7.29  & \cellcolor[rgb]{ .949,  .949,  .949}8.38 \\
    SnowflakeNet & 4.29  & 9.16  & 8.08  & 7.89  & 6.07  & 9.23  & 6.55  & 6.4   & \cellcolor[rgb]{ .949,  .949,  .949}7.21 \\
    SeedFormer & 3.85  & 9.05  & 8.06  & 7.06  & 5.21  & 8.85  & 6.05  & 5.85  & \cellcolor[rgb]{ .949,  .949,  .949}6.74 \\
    \midrule
    Point-PC & \textbf{3.73} & \textbf{8.97} & \textbf{7.79} & \textbf{6.89} & \textbf{5.01} & \textbf{8.45} & \textbf{5.82} & \textbf{5.64} & \cellcolor[rgb]{ .949,  .949,  .949}\textbf{6.62} \\
    \bottomrule
    \end{tabular}}
    \caption{Quantitative results on the PCN dataset. We report detailed results on each category and the average results under the CD-$\ell_{1}$(multiplied by 1000) metric.}
  \label{Table3}%
\end{table}%

\subsubsection{Results.}We apply the PCN dataset on Point-PC together with several SOTA methods. The CD-$\ell_{1}$ metric between the completed shapes and ground truth is reported in Table.~\ref{Table3}. Our proposed method stands out and has the best results in all the categories. In terms of average CD-$\ell_{1}$, Point-PC achieves the best score of 6.62, which illustrates that Point-PC outperforms the SOTA competitors. 

\subsection{Results on KITTI Benchmark}
\subsubsection{Data.}
The KITTI dataset \cite{Geiger2013VisionMR} consists of a series of scans from real-world environments, from which 2,401 partial car models have been extracted using 3D bounding boxes. It should be noted that these partial point clouds from the KITTI dataset do not possess corresponding complete point clouds to serve as ground truth.
We follow the standard protocol to finetune our model on ShapeNetCars \cite{Yuan2018PCNPC} and evaluate it on KITTI with metrics of Fidelity Distance and MMD(Minimal Matching Distance). We exploit KITTI to mimic a real-world situation where point clouds are sparse and irregular.
\subsubsection{Results.}
\begin{figure}[!htp]
\includegraphics[width=\linewidth]{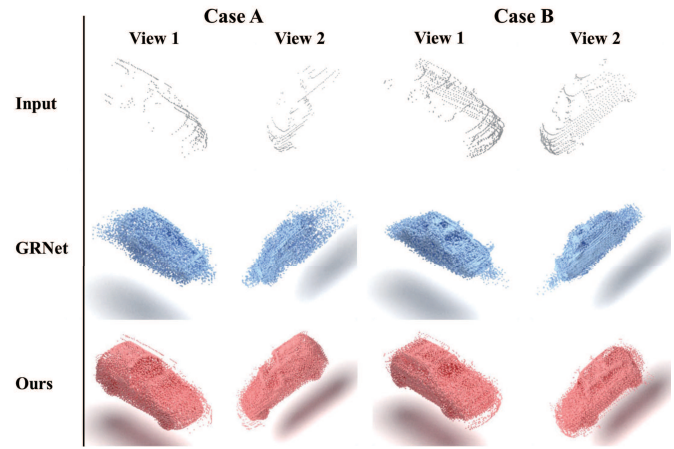}
\caption{Quantitative results on the KITTI dataset, where we present both the input partial point cloud and the predictions made by GRNet and Point-PC. For an enhanced illustration of the car's shape, two distinct views are provided for each object.} \label{Figure7}
\vspace{-.4cm}
\end{figure}

In Table.~\ref{Table4}, we present the results for both Fidelity and MMD metrics. Fidelity quantifies the mean distance from points in the input to their closest counterparts in the output, reflecting the degree to which the input has been maintained in the completed model. Meanwhile, MMD, which stands for the Chamfer Distance between the completed output and the nearest ground truth within the ShapeNetCars dataset, measures the similarity of the reconstructed model to a prototypical car shape.
Observed in Table.~\ref{Table4}, Point-PC shows better generalization ability compared with previous methods achieving a Fidelity of 0.136 and MMD of 0.509.
Qualitative results are shown in Figure.~\ref{Figure7}, which illustrates that Point-PC predicts general structures even if the input is severely sparse and proves the necessity of prior knowledge for guiding the point cloud completion in the realistic scenario.

\begin{table}[!t]
  \centering
  \small
  \resizebox{\linewidth}{!}{
    \begin{tabular}{l|cccccccccc}
    \toprule
          & FoldingNet & AtlasNet & PCN   & TopNet & MSN   & PFNet & GRNet & PoinTr & SeedFormer & Point-PC \\
    \midrule
    Fidelity $\downarrow$ & 7.467 & 1.759 & 2.235 & 5.354 & 0.434 & 1.137 & 0.816 & \textbf{0} & 0.151 & 0.136 \\
    MMD $\downarrow$ & 0.537 & 2.108 & 1.366 & 0.636 & 2.259 & 0.792 & 0.568 & 0.526 & 0.516 & \textbf{0.509} \\
    \bottomrule
    \end{tabular}}
    \caption{Quantitative results on the KITTI dataset using the metrics of Fidelity Distance and MMD (Minimal Matching Distance), where lower values indicate better performance.}
  \label{Table4}%
  \vspace{-.4cm}
\end{table}%

\subsection{Model Design Analysis}
To examine the effectiveness of our designs, we conduct detailed ablation studies.
Based on the results of settings A and B, we observed that the memory network brings a significant performance boost, demonstrating that the prior knowledge effectively compensates for the missing structural information and improves the reconstruction results. Setting C only utilizes the pre-training scheme to learn the partial shape encoder. Without the innovative prior information introduced in our paper, this improvement is limited. However, the pre-training scheme aims to help the memory network retrieve more similar prior shapes, which leads to a boost in setting D compared to setting B. The causal feature selection module extracts useful structural information that relies on the prior shapes provided by the memory network. Thus, both modules must be present together, as shown in setting E.
The results of settings D and F show improved reconstruction accuracy with the causal feature selection module. As shown in Table \ref{Table5}, the memory network improved CD-$\ell_{1}$ and F-Score@1\% by 4.84 and 0.432, respectively, while the causal feature selection module improved CD-$\ell_{1}$ and F-Score@1\% by 1.37 and 0.132, respectively. The pre-training scheme enhances the partial shape encoder by generating more similar prior shapes through the memory network. However, its effectiveness as an auxiliary design remains somewhat constrained in terms of improvement. The ablation study clearly demonstrates the effectiveness of key components in Point-PC.

\begin{table}[!t]
  \centering
  \resizebox{\linewidth}{!}{
    \begin{tabular}{l|ccc|cc}
    \toprule
    Setting & Memory Network & Pretrain Scheme & Causal Feature Selection & \cellcolor[rgb]{ .949,  .949,  .949}CD-$\ell_{1}$ $\downarrow$ & \cellcolor[rgb]{ .949,  .949,  .949}F-Score@1\% $\uparrow$ \\
    \midrule
    A     & \XSolidBrush     & \XSolidBrush     & \XSolidBrush     & \cellcolor[rgb]{ .949,  .949,  .949}15.37 & \cellcolor[rgb]{ .949,  .949,  .949}0.109 \\
    B     & \CheckmarkBold     & \XSolidBrush     & \XSolidBrush     & \cellcolor[rgb]{ .949,  .949,  .949}10.53 & \cellcolor[rgb]{ .949,  .949,  .949}0.541 \\
    C     & \XSolidBrush     & \CheckmarkBold     & \XSolidBrush     & \cellcolor[rgb]{ .949,  .949,  .949}14.01 & \cellcolor[rgb]{ .949,  .949,  .949}0.145 \\
    D     & \CheckmarkBold     & \CheckmarkBold     & \XSolidBrush     & \cellcolor[rgb]{ .949,  .949,  .949}8.22 & \cellcolor[rgb]{ .949,  .949,  .949}0.723 \\
    E     & \CheckmarkBold     & \XSolidBrush     & \CheckmarkBold     & \cellcolor[rgb]{ .949,  .949,  .949}9.16 & \cellcolor[rgb]{ .949,  .949,  .949}0.673 \\
    F     & \CheckmarkBold     & \CheckmarkBold     & \CheckmarkBold     & \cellcolor[rgb]{ .949,  .949,  .949}\textbf{6.62} & \cellcolor[rgb]{ .949,  .949,  .949}\textbf{0.759} \\
    \bottomrule
    \end{tabular}}
    \caption{Ablation study on the PCN dataset. We investigate different designs including the Memory Network, the pre-training scheme, and the causal feature selection module.}
  \label{Table5}%
\end{table}%

\begin{table}[t]
  \centering
  \resizebox{\linewidth}{!}{
    \begin{tabular}{l|cccccc}
    \toprule
    Shape Prior Num. & 0     & 1     & 2     & 3     & 4     & 5 \\
    \midrule
    CD-AVG & 15.37 & 9.7   & 7.64  & \textbf{6.62}   & 7.57  & 10.12 \\
    F-Score@1\% & 0.109 & 0.647 & 0.732 & \textbf{0.759} & 0.740 & 0.556 \\
    \bottomrule
    \end{tabular}}
    \caption{Ablation study on the number of the shape priors.}
  \label{Table6}
\end{table}%
\begin{table}[ht]
  \centering
  \resizebox{\linewidth}{!}{
    \begin{tabular}{l|cccccc}
    \toprule
    Threshold & 0     & 0.5   & 1     & 1.5   & 2     & 2.5 \\
    \midrule
    CD-AVG & 10.94 & 10.87 & 7.29  & \textbf{6.62}   & 7.06  & 9.23 \\
    F-Score@1\% & 0.531 & 0.534 & 0.745 & \textbf{0.759} & 0.752 & 0.65 \\
    \bottomrule
    \end{tabular}}
    \caption{Ablation study on the similarity threshold $\delta$(multiplied by 0.001).}
  \label{Table7}
\end{table}%

We also present the ablation experiment on the number of shape priors in Table \ref{Table6}. According to the metrics, the performance of Point-PC tends to go upward as the number of shape priors increases from zero but goes downward immediately after the shape priors exceed a certain amount. This is mainly because the shape priors explicitly make up the deficiency of the partial input representing the geometry of missing parts, yet excessive shape priors introduce too many confounding structures which is beyond the capability of the causal feature selection module.
Particularly, Point-PC achieves the best score using 3 shape priors.

In Section 3.1, we adopt the Chamfer Distance threshold $\delta$ to determine whether the partial point cloud and the "key-value" pair constitute a valid positive match. We set $\delta$ within a certain range, and report the results in Table \ref{Table7}. Point-PC achieves the best metrics of average CD-$\ell_{1}$ and F-Score@1\% when the threshold $\delta$ is set to 1.5.

\subsection{Detailed results on ShapeNet-55}
A comprehensive performance analysis of FoldingNet, PCN, TopNet, PFNet, GRNet, and our proposed method on ShapeNet-55 is presented in Table \ref{Table9}. Each row in the table represents a distinct object category, and we have assessed the performance of each method across three different settings: simple, moderate, and hard.

\subsection{Detailed results on ShapeNet-34}
We provide a comprehensive report of the results obtained for novel objects across 21 categories in ShapeNet-34, as shown in Table \ref{Table10}. Each row in the table represents a specific object category, and we have evaluated each method under three different settings: simple, moderate, and hard.

\subsection{Complexity Analysis}
Table \ref{Table11} shows a comprehensive examination of the complexity of our method, encompassing the computation cost(FLOPs) and the parameter counts, in comparison to five other methods. Moreover, we present the average Chamfer distances of all categories in ShapeNet-55 as well as unseen categories in ShapeNet-34 as references. Our method demonstrates superior performance while maintaining relatively low FLOPs and params compared to the other methods, as depicted in the table. This highlights the favorable balance our method achieves between cost and performance.

\begin{table}[htbp]
  \centering
  \resizebox{\linewidth}{!}{
    \begin{tabular}{l|ccccccc}
    \toprule
    Models & FoldingNet & PCN   & TopNet & PFNet & GRNet & PoinTr & Point-PC \\
    \midrule
    Params(M) & 2.3   & 5.04  & 5.76  & 73.05 & 73.15 & 30.9  & 10.29 \\
    FLOPs(G)  & 27.58 & 15.25 & 6.72  & 4.96  & 40.44 & 10.41 & 14.88 \\
    CD55  & 3.12  & 2.66  & 2.91  & 5.22  & 1.97  & 1.07  & 0.83 \\
    CD34  & 3.62  & 3.85  & 3.5   & 8.16  & 2.99  & 2.05  & 1.19 \\
    \bottomrule
    \end{tabular}}%
  \caption{We analyze the complexity of our method and five other methods, reporting the parameter counts(Params) and computation cost (FLOPs). Additionally, we have calculated the average CD-$\ell_{2}$ for ShapeNet-55 (CD55) and ShapeNet-34 (CD34) to establish a basis for comparison.}
  \label{Table11}
\end{table}%

\begin{table*}[!t]
  \centering
  \resizebox{\textwidth}{!}{
    \begin{tabular}{|l|ccc|ccc|ccc|ccc|ccc|ccc|}
    \toprule
    \multirow{2}[4]{*}{CD-$\ell_{2}$(× 1000)} & \multicolumn{3}{c|}{TopNet} & \multicolumn{3}{c|}{PCN} & \multicolumn{3}{c|}{GRNet} & \multicolumn{3}{c|}{PoinTr} & \multicolumn{3}{c|}{SeedFormer} & \multicolumn{3}{c|}{Point-PC} \\
\cmidrule{2-19}          & S.    & M.    & H.    & S.    & M.    & H.    & S.    & M.    & H.    & S.    & M.    & H.    & S.    & M.    & H.    & S.    & M.    & H. \\
    \midrule
    bag   & 2.08  & 1.95  & 4.36  & 2.48  & 2.46  & 3.94  & 1.47  & 1.88  & 3.45  & 0.96  & 1.34  & 2.08  & 0.49  & 0.82  & 1.45  & 0.61  & 1.15  & 1.9 \\
    basket & 2.46  & 2.11  & 5.18  & 2.79  & 2.51  & 4.78  & 1.78  & 1.94  & 4.18  & 1.04  & 1.4   & 2.9   & 0.6   & 0.85  & 1.98  & 0.64  & 1.12  & 2.23 \\
    birdhouse & 3.17  & 2.97  & 5.89  & 3.53  & 3.47  & 5.31  & 1.89  & 2.34  & 5.16  & 1.22  & 1.79  & 3.45  & 0.72  & 1.19  & 2.31  & 0.89  & 1.22  & 2.32 \\
    bowl  & 2.46  & 2.16  & 4.84  & 2.66  & 2.35  & 3.97  & 1.77  & 1.97  & 3.9   & 1.05  & 1.32  & 2.4   & 0.6   & 0.77  & 1.5   & 0.53  & 1.06  & 1.73 \\
    camera & 4.24  & 4.43  & 8.11  & 4.84  & 5.3   & 8.03  & 2.31  & 3.38  & 7.2   & 1.63  & 2.67  & 4.97  & 0.89  & 1.77  & 3.75  & 0.8   & 1.4   & 2.32 \\
    can   & 2.02  & 1.7   & 5.82  & 1.95  & 1.89  & 5.21  & 1.53  & 1.8   & 3.08  & 0.8   & 1.17  & 2.85  & 0.56  & 0.89  & 1.57  & 0.58  & 0.81  & 1.75 \\
    cap   & 4.68  & 4.23  & 9.7   & 7.21  & 7.4   & 10.94 & 3.29  & 4.87  & 13.02 & 1.4   & 2.74  & 8.35  & 0.5   & 1.34  & 5.19  & 0.6   & 1.25  & 3.81 \\
    keyboard & 0.79  & 0.77  & 1.55  & 1.07  & 1     & 1.23  & 0.73  & 0.77  & 1.11  & 0.43  & 0.45  & 0.63  & 0.32  & 0.41  & 0.6   & 0.29  & 0.44  & 0.66 \\
    dishwasher & 2.51  & 1.77  & 4.72  & 2.45  & 2.09  & 3.53  & 1.79  & 1.7   & 3.27  & 0.93  & 1.05  & 2.04  & 0.63  & 0.78  & 1.44  & 0.53  & 0.81  & 1.74 \\
    earphone & 5.33  & 4.83  & 11.67 & 7.88  & 6.59  & 16.53 & 4.29  & 4.16  & 10.3  & 2.03  & 5.1   & 10.69 & 1.18  & 2.78  & 6.71  & 0.95  & 1.24  & 4.82 \\
    helmet & 4.89  & 4.86  & 8.73  & 6.15  & 6.41  & 9.16  & 3.06  & 4.38  & 10.27 & 1.86  & 3.3   & 6.96  & 1.1   & 2.27  & 4.78  & 0.79  & 0.68  & 2.56 \\
    mailbox & 2.35  & 2.2   & 4.91  & 2.74  & 2.68  & 4.31  & 1.2   & 1.9   & 4.33  & 1.03  & 1.47  & 3.34  & 0.56  & 0.99  & 2.06  & 0.44  & 1.01  & 2.45 \\
    microphone & 3.03  & 3.2   & 7.15  & 4.36  & 4.65  & 8.46  & 2.29  & 3.23  & 8.4   & 1.25  & 2.27  & 5.47  & 0.8   & 1.61  & 4.21  & 0.64  & 0.61  & 3.59 \\
    microwaves & 2.67  & 2.12  & 5.41  & 2.59  & 2.35  & 4.47  & 1.74  & 1.81  & 3.82  & 1.01  & 1.18  & 2.14  & 0.64  & 0.83  & 1.69  & 0.45  & 0.91  & 1.92 \\
    pillow & 2.08  & 2.05  & 4.01  & 2.09  & 2.16  & 3.54  & 1.43  & 1.69  & 3.43  & 0.92  & 1.24  & 2.39  & 0.43  & 0.66  & 1.45  & 0.66  & 0.79  & 2.26 \\
    printer & 2.9   & 2.96  & 6.07  & 3.28  & 3.6   & 5.56  & 1.82  & 2.41  & 5.09  & 1.18  & 1.76  & 3.1   & 0.69  & 1.25  & 2.33  & 0.74  & 1.37  & 2.44 \\
    remote & 0.89  & 0.89  & 2.28  & 0.95  & 1.08  & 1.58  & 0.82  & 1.02  & 1.29  & 0.44  & 0.58  & 0.78  & 0.27  & 0.42  & 0.61  & 0.3   & 0.39  & 0.54 \\
    rocket & 1.14  & 0.96  & 2.03  & 1.39  & 1.22  & 2.01  & 0.97  & 0.79  & 1.6   & o.39  & o.72  & 1.39  & 0.28  & 0.51  & 1.02  & 0.32  & 0.56  & 1.56 \\
    skatcboard & 1.23  & 1.2   & 2.01  & 1.97  & 1.78  & 2.45  & 0.93  & 1.07  & 1.83  & 0.52  & 0.8   & 1.31  & 0.35  & 0.56  & 0.92  & 0.34  & 0.44  & 0.94 \\
    tower & 2.2   & 2.17  & 5.47  & 2.37  & 2.4   & 4.35  & 1.35  & 1.8   & 3.85  & 0.82  & 1.35  & 2.48  & 0.51  & 0.92  & 1.87  & 0.64  & 0.93  & 1.8 \\
    washer & 2.63  & 2.14  & 6.57  & 2.77  & 2.52  & 4.64  & 1.83  & 1.97  & 5.28  & 1.04  & 1.39  & 2.73  & 0.61  & 0.87  & 1.94  & 0.47  & 1.11  & 1.86 \\
    \midrule
    mean  & 2.65  & 2.46  & 5.52  & 3.22  & 3.13  & 5.43  & 1.84  & 2.23  & 4.95  & 1.05  & 1.67  & 3.45  & 0.61  & 1.07  & 2.35  & 0.57  & 0.92  & 2.09 \\
    \bottomrule
    \end{tabular}}%
  \caption{Detailed quantitative results for the novel categories on ShapeNet-34. S., M. and H. represent the simple, moderate and hard settings.}
  \label{Table10}
\end{table*}%

\begin{table*}[t]
  \centering
  \resizebox{\textwidth}{!}{
    \begin{tabular}{|l|ccc|ccc|ccc|ccc|ccc|ccc|}
    \toprule
    \multirow{2}[4]{*}{CD-$\ell_{2}$(× 1000)} & \multicolumn{3}{c|}{TopNet} & \multicolumn{3}{c|}{PCN} & \multicolumn{3}{c|}{GRNet} & \multicolumn{3}{c|}{PoinTr} & \multicolumn{3}{c|}{SeedFormer} & \multicolumn{3}{c|}{Point-PC} \\
\cmidrule{2-19}          & S.    & M.    & H.    & S.    & M.    & H.    & S.    & M.    & H.    & S.    & M.    & H.    & S.    & M.    & H.    & S.    & M.    & H. \\
    \midrule
    airplane  & 1.02  & 0.99  & 1.48  & 0.9   & 0.89  & 1.32  & 0.87  & 0.87  & 1.27  & 0.27  & 0.38  & 0.69  & 0.23  & 0.35  & 0.61  & 0.28  & 0.31  & 0.54 \\
    trash bin & 2.51  & 2.32  & 5.03  & 2.16  & 2.18  & 5.15  & 1.69  & 2.01  & 3.48  & 0.8   & 1.15  & 2.15  & 0.73  & 1.08  & 1.94  & 0.65  & 0.93  & 1.74 \\
    bag   & 2.36  & 2.23  & 4.21  & 2.11  & 2.04  & 4.44  & 1.41  & 1.7   & 2.97  & 0.53  & 0.74  & 1.51  & 0.43  & 0.67  & 1.28  & 0.41  & 0.78  & 1.22 \\
    basket & 2.62  & 2.43  & 5.71  & 2.21  & 2.1   & 4.55  & 1.65  & 1.84  & 3.15  & 0.73  & 0.88  & 1.82  & 0.65  & 0.83  & 1.54  & 0.55  & 0.84  & 1.32 \\
    bathtub & 2.49  & 2.25  & 4.33  & 2.11  & 2.09  & 3.94  & 1.46  & 1.73  & 2.73  & 0.64  & 0.94  & 1.68  & 0.52  & 0.82  & 1.45  & 0.47  & 0.81  & 1.39 \\
    bed   & 3.13  & 3.1   & 5.71  & 2.86  & 3.07  & 5.54  & 1.64  & 2.03  & 3.7   & 0.76  & 1.1   & 2.26  & 0.63  & 0.91  & 1.89  & 0.54  & 0.81  & 1.94 \\
    bench  & 1.56  & 1.39  & 2.4   & 1.31  & 1.24  & 2.14  & 1.03  & 1.09  & 1.71  & 0.38  & 0.52  & 0.94  & 0.32  & 0.42  & 0.84  & 0.33  & 0.55  & 0.94 \\
    birdhouse  & 3.73  & 3.98  & 6.8   & 3.29  & 3.53  & 6.69  & 1.87  & 2.4   & 4.71  & 0.98  & 1.49  & 3.13  & 0.76  & 1.3   & 2.46  & 0.66  & 1.17  & 2.29 \\
    bookshelf & 3.11  & 2.87  & 4.87  & 2.7   & 2.7   & 4.61  & 1.42  & 1.71  & 2.78  & 0.71  & 1.06  & 1.93  & 0.57  & 0.84  & 1.57  & 0.51  & 0.97  & 1.68 \\
    bottle & 1.56  & 1.66  & 4.02  & 1.25  & 1.43  & 4.61  & 1.05  & 1.44  & 2.67  & 0.37  & 0.74  & 1.5   & 0.31  & 0.63  & 1.21  & 0.34  & 0.58  & 1.06 \\
    bowl  & 2.33  & 1.98  & 4.82  & 2.05  & 1.83  & 3.66  & 1.6   & 1.77  & 2.99  & 0.68  & 0.78  & 1.44  & 0.56  & 0.65  & 1.18  & 0.47  & 0.86  & 1.24 \\
    bus   & 1.32  & 1.21  & 2.29  & 1.2   & 1.14  & 2.08  & 1.06  & 1.16  & 1.48  & 0.42  & 0.55  & 0.79  & 0.42  & 0.55  & 0.73  & 0.39  & 0.62  & 0.77 \\
    cabinet  & 1.91  & 1.65  & 3.36  & 1.6   & 1.49  & 3.47  & 1.27  & 1.41  & 2.09  & 0.55  & 0.66  & 1.16  & 0.57  & 0.69  & 1.05  & 0.57  & 0.68  & 1.03 \\
    camera  & 4.75  & 4.98  & 9.24  & 4.05  & 4.54  & 8.27  & 2.14  & 3.15  & 6.09  & 1.1   & 2.03  & 4.34  & 0.83  & 1.68  & 3.45  & 0.75  & 1.29  & 2.24 \\
    can   & 2.67  & 2.4   & 5.5   & 2.02  & 2.28  & 6.48  & 1.58  & 2.11  & 3.81  & 0.68  & 1.19  & 2.14  & 0.58  & 1.03  & 1.79  & 0.44  & 1.18  & 1.76 \\
    cap   & 3     & 2.69  & 5.59  & 1.82  & 1.76  & 4.2   & 1.17  & 1.37  & 3.05  & 0.46  & 0.62  & 1.64  & 0.33  & 0.45  & 1.18  & 0.41  & 0.69  & 1.41 \\
    car   & 1.71  & 1.65  & 3.17  & 1.48  & 1.47  & 2.6   & 1.29  & 148   & 2.14  & 0.64  & 0.86  & 1.25  & 0.65  & 0.86  & 1.17  & 0.55  & 0.91  & 1.05 \\
    cellphone & 1.01  & 0.96  & 1.8   & 0.8   & 0.79  & 1.71  & 0.82  & 0.91  & 1.18  & 0.32  & 0.39  & 0.6   & 0.31  & 0.4   & 0.54  & 0.37  & 0.34  & 0.59 \\
    chair & 1.97  & 2.04  & 3.59  & 1.7   & 1.81  & 3.34  & 1.24  & 1.56  & 2.73  & 0.49  & 0.74  & 1.63  & 0.41  & 0.65  & 1.38  & 0.32  & 0.51  & 0.98 \\
    clock & 2.48  & 2.16  & 4.03  & 2.1   & 2.01  & 3.98  & 1.46  & 1.66  & 2.67  & 0.62  & 0.84  & 1.65  & 0.53  & 0.74  & 1.35  & 0.39  & 0.66  & 1.25 \\
    keyboard & 0.88  & 0.83  & 1.5   & 0.82  & 0.82  & 1.04  & 0.74  & 0.81  & 1.09  & 0.3   & 0.39  & 0.45  & 0.28  & 0.36  & 0.45  & 0.32  & 0.39  & 0.44 \\
    dishwasher & 2.43  & 1.74  & 4.64  & 1.93  & 1.66  & 4.39  & 1.43  & 1.59  & 2.53  & 0.55  & 0.69  & 1.42  & 0.56  & 0.69  & 1.3   & 0.44  & 0.58  & 1.27 \\
    display & 1.84  & 1.85  & 3.48  & 1.56  & 1.66  & 3.26  & 1.13  & 1.38  & 2.29  & 0.48  & 0.67  & 1.33  & 0.39  & 0.59  & 1.1   & 0.29  & 0.41  & 0.95 \\
    earphone  & 4.36  & 4.47  & 8.36  & 3.13  & 2.94  & 7.56  & 1.78  & 2.18  & 5.33  & 0.81  & 1.38  & 3.78  & 0.64  & 1.04  & 2.75  & 0.76  & 0.88  & 1.79 \\
    faucet & 3.61  & 3.59  & 7.25  & 3.21  & 3.48  & 7.52  & 1.81  & 2.32  & 4.91  & 0.71  & 1.42  & 3.49  & 0.55  & 1.15  & 2.63  & 0.55  & 1.02  & 1.94 \\
    filecabinet & 2.41  & 2.12  & 4.12  & 2.02  & 1.97  & 4.14  & 1.46  & 1.71  & 2.89  & 0.63  & 0.84  & 1.69  & 0.63  & 0.84  & 1.49  & 0.68  & 0.97  & 1.35 \\
    guitar & 0.57  & 0.47  & 1.42  & 0.42  & 0.38  & 1.23  & 0.44  & 0.48  & 0.76  & 0.14  & 0.21  & 0.42  & 0.13  & 0.19  & 0.32  & 0.27  & 0.24  & 0.36 \\
    helmet & 4.36  & 4.55  & 7.73  & 3.76  & 4.18  & 7.53  & 2.33  & 3.18  & 6.03  & 0.99  & 1.93  & 4.22  & 0.79  & 1.52  & 3.61  & 0.68  & 1.14  & 2.22 \\
    jar   & 3.03  & 3.17  & 7.03  & 2.57  & 2.82  & 6.01  & 1.72  & 2.37  & 4.37  & 0.77  & 1.33  & 2.87  & 0.63  & 1.13  & 2.36  & 0.57  & 1.19  & 2.08 \\
    knife & 0.84  & 0.68  & 1.44  & 0.94  & 0.62  & 1.37  & 0.72  & 0.66  & 0.96  & 0.2   & 0.33  & 0.56  & 0.15  & 0.28  & 0.45  & 0.15  & 0.33  & 0.42 \\
    lamp  & 3.03  & 3.39  & 8.15  & 3.1   & 3.45  & 7.02  & 1.68  & 2.43  & 5.17  & 0.64  & 1.4   & 3.58  & 0.45  & 1.06  & 2.67  & 0.2   & 0.98  & 1.95 \\
    laptop & 0.8   & 0.85  & 1.66  & 0.75  & 0.79  & 1.59  & 0.83  & 0.87  & 1.28  & 0.32  & 0.34  & 0.6   & 0.32  & 0.37  & 0.55  & 0.35  & 0.37  & 0.68 \\
    loudspeaker & 3.1   & 2.76  & 5.32  & 2.5   & 2.45  & 5.08  & 1.75  & 2.08  & 3.45  & 0.78  & 1.16  & 2.17  & 0.67  & 1.01  & 1.8   & 0.52  & 1.29  & 1.95 \\
    mailbox  & 2.16  & 2.1   & 5.1   & 1.66  & 1.74  & 5.18  & 1.15  & 1.59  & 3.42  & 0.39  & 0.78  & 2.56  & 0.3   & 0.67  & 2.04  & 0.29  & 0.43  & 1.35 \\
    microphone & 2.83  & 3.49  & 6.87  & 3.4   & 3.9   & 8.52  & 2.09  & 2.76  & 5.7   & 0.7   & 1.66  & 4.48  & 0.62  & 1.61  & 3.66  & 0.48  & 1.21  & 2.85 \\
    microwaves & 2.65  & 2.15  & 5.07  & 2.2   & 2.01  & 4.65  & 1.51  & 1.72  & 2.76  & 0.67  & 0.83  & 1.82  & 0.63  & 0.79  & 1.47  & 0.41  & 0.59  & 1.38 \\
    motorbike & 2.29  & 2.25  & 3.54  & 2.03  & 2.01  & 3.13  & 1.38  & 1.52  & 2.26  & 0.75  & 1.1   & 1.92  & 0.68  & 0.96  & 1.44  & 0.65  & 0.93  & 1.48 \\
    mug   & 2.89  & 2.56  & 5.43  & 2.45  & 2.48  & 5.17  & 1.75  & 2.16  & 3.79  & 0.91  & 1.17  & 2.35  & 0.79  & 1.03  & 2.06  & 0.61  & 0.81  & 1.72 \\
    piano & 2.99  & 2.89  & 5.64  & 2.64  & 2.74  & 4.83  & 1.53  & 1.82  & 3.21  & 0.76  & 1.06  & 2.23  & 0.62  & 0.87  & 1.79  & 0.53  & 0.77  & 1.7 \\
    pillow & 2.31  & 2.26  & 4.19  & 1.85  & 1.81  & 3.68  & 1.42  & 1.67  & 3.04  & 0.61  & 0.82  & 1.56  & 0.48  & 0.75  & 1.41  & 0.32  & 0.73  & 1.36 \\
    pistol & 1.5   & 1.3   & 2.62  & 1.25  & 1.17  & 2.65  & 1.1   & 1.06  & 1.76  & 0.43  & 0.66  & 1.3   & 0.37  & 0.56  & 0.96  & 0.29  & 0.54  & 0.79 \\
    flowerpot & 3.61  & 3.45  & 6.28  & 3.32  & 3.39  & 6.04  & 2.02  & 2.48  & 4.19  & 1.01  & 1.51  & 2.77  & 0.93  & 1.3   & 2.32  & 1.09  & 1.36  & 2.53 \\
    printer & 3.04  & 3.19  & 5.84  & 2.9   & 3.19  & 5.84  & 1.56  & 2.38  & 4.24  & 0.73  & 1.21  & 2.47  & 0.58  & 1.11  & 2.13  & 0.54  & 0.83  & 2.18 \\
    remote & 1.14  & 1.17  & 2.16  & 0.99  & 0.97  & 2.04  & 0.89  & 1.05  & 1.29  & 0.36  & 0.53  & 0.71  & 0.29  & 0.46  & 0.62  & 0.26  & 0.46  & 0.68 \\
    rifle & 0.98  & 0.86  & 1.46  & 0.98  & 0.8   & 1.31  & 0.83  & 0.77  & 1.16  & 0.3   & 0.45  & 0.79  & 0.27  & 0.41  & 0.66  & 0.24  & 0.44  & 0.63 \\
    rocket & 1.04  & 1     & 1.93  & 1.05  & 104   & 1.87  & 0.78  & 0.92  & 1.44  & 0.23  & 0.48  & 0.99  & 0.21  & 0.46  & 0.83  & 0.26  & 0.55  & 0.75 \\
    skateboard & 1.08  & 1.05  & 1.84  & 1.04  & 0.94  & 1.68  & 0.82  & 0.87  & 1.24  & 0.28  & 0.38  & 0.62  & 0.23  & 0.32  & 0.62  & 0.21  & 0.31  & 0.59 \\
    sofa  & 1.93  & 1.76  & 3.39  & 1.65  & 1.61  & 2.92  & 1.35  & 1.45  & 2.32  & 0.56  & 0.67  & 1.14  & 0.5   & 0.62  & 1.02  & 0.42  & 0.58  & 0.96 \\
    stove & 2.44  & 2.16  & 4.84  & 2.07  & 2.02  & 4.72  & 1.46  & 1.72  & 3.22  & 0.63  & 0.92  & 1.73  & 0.59  & 0.87  & 1.49  & 0.46  & 0.56  & 1.23 \\
    table  & 1.78  & 1.65  & 3.21  & 1.56  & 1.5   & 3.36  & 1.15  & 1.33  & 2.33  & 0.46  & 0.64  & 1.31  & 0.41  & 0.58  & 1.18  & 0.36  & 0.51  & 1.09 \\
    telephone  & 1.02  & 0.95  & 1.78  & 0.8   & 0.8   & 1.67  & 0.81  & 0.89  & 1.18  & 0.31  & 0.38  & 0.59  & 0.31  & 0.39  & 0.55  & 0.32  & 0.35  & 0.54 \\
    tower  & 2.15  & 2.05  & 4.51  & 1.91  & 1.97  & 4.47  & 1.26  & 1.69  & 3.06  & 0.55  & 0.9   & 1.95  & 0.47  & 0.84  & 1.65  & 0.35  & 0.67  & 1.31 \\
    train & 1.59  & 1.44  & 2.51  & 1.5   & 1.41  & 2.37  & 1.09  & 1.14  & 1.61  & 0.5   & 0.7   & 1.12  & 0.51  & 0.66  & 1.01  & 0.59  & 0.63  & 1.05 \\
    watercraft  & 1.53  & 1.42  & 2.67  & 1.46  & 1.39  & 2.4   & 1.09  & 1.12  & 1.65  & 0.41  & 0.62  & 1.07  & 0.35  & 0.56  & 0.92  & 0.32  & 0.55  & 0.89 \\
    washer & 2.92  & 2.53  & 6.53  & 2.42  & 2.31  & 6.08  & 1.72  & 2.05  & 4.19  & 0.75  & 1.06  & 2.44  & 0.64  & 0.91  & 2.04  & 0.38  & 0.73  & 1.85 \\
    \midrule
    mean  & 2.26  & 2.17  & 4.31  & 1.96  & 1.98  & 4.09  & 1.35  & 1.63  & 2.86  & 0.58  & 0.88  & 1.8   & 0.5   & 0.77  & 1.49  & 0.45  & 0.72  & 1.32 \\
    \bottomrule
    \end{tabular}}%
    \caption{Detailed quantitative results on the ShapeNet-55 dataset. S., M., and H. represent simple, moderate, and hard settings.}
  \label{Table9}
\end{table*}%

\subsection{Implementation Details}
We employ a geometry-aware transformer encoder \cite{Yu2021PoinTrDP} to extract the point cloud features. In all of our experiments, 
we use a uniform configuration for all transformer encoders, utilizing 6 head attention and 8 block depth, with a hidden dimension of 384. In terms of geometry-aware modules, we set the k value of the kNN algorithm to 8 and 16 for those in the DGCNN \cite{Wang2018DynamicGC} feature extractor.
We set the number of partial shape patches that contain 2048 points to 32. On the ShapeNet-55/34 dataset, we set the complete shape of 8192 points to 128 neighboring patches, while 256 complete shape patches of 16384 points on PCN. The threshold $\delta$ mentioned above is set to 0.0015 and we finally select top-3 shape priors from the memory.
During the evaluation of our method on the ShapeNet-55 and ShapeNet-34/21 benchmarks, we set 8 fixed viewpoints and select 2048, 4096, or 6144 points (25\%, 50\%, or 75\% of the entire point cloud) for ease of evaluation. Consequently, we classify the test samples into three difficulty levels, namely simple, moderate, and hard. Furthermore, we report the overall performance (Avg) by taking the average of the performance across all three difficulty levels.
We have implemented our networks using PyTorch and trained our models using the AdamW optimizer \cite{Loshchilov2017FixingWD}. The initial learning rate is set at 5e-4 and decays by 0.76 every 20 epochs. With a batch size of 64, we trained the model for 200 epochs using two NVIDIA RTX 3090Ti GPUs.

\section{Conclusion}
In this paper, We emphasize that the existing methods have two drawbacks: 1) incomplete point cloud is inadequate to provide missing structural information. The reason is that the information does not exist; 2) The global features are spread out in coarse-to-fine up-sampling operations, limiting their ability to capture geometric information. To address these issues, we design the memory network, which can search for the relevant complete shapes (prior shapes) corresponding to the partial point cloud input. These prior shapes can make up for the missing structural information and then guide the generative model to fill in more accurate details. Note that only part of the structure of these prior shapes is helpful for the generation of missing structure information, while the remaining structure information can be regarded as redundant information. Therefore, we utilize causal inference to mitigate the confounding effect of shape priors and to encourage the decoder to pay more attention to causal selected features that truly contribute to the accuracy of completion.
To our best knowledge, this is the first work to introduce a causal graph into the point cloud completion task, which effectively filters shape information from previous shapes and preserves missing shape information to improve the integrity and ultimate performance of the fused representation. Comprehensive experiments show the effectiveness and superiority of Point-PC compared to state-of-the-art competitors.

\bibliographystyle{ieeetr}
\bibliography{reference}

\vfill

\end{document}